\documentclass[11pt]{article}
\usepackage{coling2018}
\usepackage{times}
\usepackage{latexsym}
\usepackage{color, soul}
\usepackage{times}
\usepackage{xcolor}
\usepackage{soul}
\usepackage[utf8]{inputenc}
\usepackage[small]{caption}
\usepackage{url}
\usepackage{amssymb}
\usepackage{amsmath}
\usepackage{amsfonts}
\usepackage{floatrow}
\usepackage{bm}
\usepackage{amssymb}
\usepackage{graphicx,bm}
\usepackage[shortlabels]{enumitem}
\usepackage{multirow,multicol}
\usepackage{bigstrut}
\usepackage{MnSymbol}
\usepackage{algorithm}
\usepackage{pifont}
\usepackage{microtype}
\usepackage{makecell}
\usepackage{subcaption}
\newcommand{\cmark}{\text{\ding{51}}}

\newfloatcommand{capbtabbox}{table}[][\FBwidth]
\definecolor{mygreen}{RGB}{0, 100, 0}
\title{CASCADE: Contextual Sarcasm Detection in Online Discussion Forums\vspace{0.5cm}}

\author{Devamanyu Hazarika \\
  \small{School of Computing,}\\
  \small{National University of Singapore}\\
  {\footnotesize \tt hazarika@comp.nus.edu.sg} \\\And  
  Soujanya Poria \\
  \small{Artificial Intelligence Initiative,} \\
  \small{A*STAR, Singapore}\\
  {\small \tt sporia@ihpc.a-star.edu.sg} \\\And  
  Sruthi Gorantla \\
  \small{Computer Science \& Automation}\\
  \small{Indian Institute of Science, Bangalore}\\
  {\small \tt gorantlas@iisc.ac.in } \\ \AND
  Erik Cambria \\
  \small{School of Computer Science and}\\
  \small{Engineering, NTU, Singapore}\\
  {\footnotesize \tt cambria@ntu.edu.sg} \\\And  
  Roger Zimmermann \\
  \small{School of Computing,}\\
  \small{National University of Singapore}\\ 
  {\small \tt rogerz@comp.nus.edu.sg} \\ \And
  Rada Mihalcea \\
  \small{Computer Science \& Engineering}\\
  \small{University of Michigan, Ann Arbor}\\
  {\small \tt mihalcea@umich.edu} \\
}

\date{}

\begin{document}
\maketitle
\begin{abstract}
The literature in automated sarcasm detection has mainly focused on lexical, syntactic and semantic-level analysis of text. However, a sarcastic sentence can be expressed with contextual presumptions, background and commonsense knowledge. 
In this paper, we propose CASCADE (a ContextuAl SarCasm DEtector) that adopts a hybrid approach of both content and context-driven modeling for sarcasm detection in online social media discussions. For the latter, CASCADE aims at extracting contextual information from the discourse of a discussion thread. Also, since the sarcastic nature and form of expression can vary from person to person, CASCADE utilizes user embeddings that encode
stylometric and personality features of the users. 
When used along with content-based feature extractors such as Convolutional Neural Networks (CNNs), we see a significant boost in the classification performance on a large Reddit corpus.
\end{abstract}

\section{Introduction} \label{Intro}

Sarcasm is a linguistic tool that uses irony to express contempt. Its figurative nature poses a great challenge for affective systems performing sentiment analysis. Previous research in automated sarcasm detection has primarily focused on lexical, pragmatic cues found in sentences~\cite{kreuz2007lexical}. Interjections, punctuations, sentimental shifts, etc., have been considered as major indicators of sarcasm~\cite{joshi2017automatic}. When such lexical cues are present in sentences, sarcasm detection can achieve high accuracy. However, sarcasm is also expressed implicitly, i.e., without the use of any explicit lexical cues. Such use of sarcasm also relies on the context which involves the presumption of commonsense and background knowledge of an event. When it comes to detecting sarcasm in a discussion forum, it may not only require understanding the context of the previous comments but also need necessary external background knowledge about the topic of discussion.
The usage of slangs and informal language also diminishes the reliance on lexical cues. This particular type of sarcasm is tough to detect~\cite{poria2016deeper}.

Contextual dependencies for sarcasm can take many forms. As an example, a sarcastic post from Reddit\footnote{\protect\url{https://www.reddit.com/}}, \textit{``I'm sure Hillary would've done that, lmao.''} requires background knowledge about the event, i.e., Hillary Clinton's action at the time the post was made. Similarly, sarcastic posts like \textit{``But atheism, yeah *that's* a religion!''} requires the knowledge that topics like \textit{atheism} often contain argumentative discussions and are more prone towards sarcasm. 

In this work, we attempt the task of sarcasm detection in online discussion forums. Particularly, we propose 
a hybrid network, named CASCADE, that utilizes both \textit{content} and \textit{contextual-}information required for sarcasm detection. It starts by processing contextual information in two ways. First, it performs user profiling to create user embeddings that capture indicative behavioral traits for sarcasm. Recent findings suggest that such modeling of the user and their preferences, is highly effective for the given task~\cite{amir2016modelling}. It makes use of users' historical posts to model their writing style (stylometry) and personality indicators, which are then fused into comprehensive user embeddings using a multi-view fusion approach, Canonical Correlation Analysis (CCA). Second, it extracts contextual information from the discourse of comments in the discussion forums. This is done by document modeling of these consolidated comments belonging to the same forum. We hypothesize that these discourse features would give the important contextual information, background cues along with topical information required for detecting sarcasm.

After the contextual modeling phase, CASCADE is provided with a comment for sarcasm detection. It performs content-modeling using a Convolutional Neural Network (CNN) to extract its syntactic features. This CNN representation is then concatenated with the relevant user embedding and discourse features to get the final representation which is used for classification. The overall contribution of this work can be summarized as:
\begin{itemize}[leftmargin=*,noitemsep]
    \item We propose a novel hybrid sarcasm detector, CASCADE that models content and contextual information.
      \item We model stylometric and personality details of users along with discourse features of discussion forums to learn informative contextual representations. Experiments on a large Reddit corpus, SARC, demonstrate significant performance improvement over state-of-the-art automated sarcasm detectors. 
\end{itemize}
In the remaining paper, Section~\ref{sec:related} compares our model to related works; Section~\ref{sec:method} provides the task description and proposed approach; here, Section~\ref{sec:context} explains the process of learning contextual features comprising user embeddings and discourse features; Section~\ref{sec:prediction} presents the hybrid prediction model followed by experimentation details and result analysis in Section~\ref{sec:experiments}; finally, Section~\ref{sec:conclusion} draws conclusion.


\section{Related Work} \label{sec:related}
Automated sarcasm detection is a relatively recent field of research. The previous works in the literature can be largely classified into two categories, content and context-based sarcasm detection models.

\paragraph{Content-based: } These networks model the problem of sarcasm detection as a standard classification task and try to find lexical and pragmatic indicators to identify sarcasm. Numerous works have taken this path and presented innovative ways to unearth interesting cues for sarcasm.~\newcite{tepperman2006yeah} investigate sarcasm detection in spoken dialogue systems using prosodic and spectral cues. ~\newcite{carvalho2009clues} use linguistic features like positive predicates, interjections and gestural clues such as emoticons, quotation marks, etc.~\newcite{davidov2010semi},~\newcite{tsur2010icwsm} use syntactic patterns to construct classifiers.~\newcite{gonzalez2011identifying} also study the use of emoticons, mainly amongst tweets.~\newcite{riloff2013sarcasm} assert sarcasm to be a contrast to positive sentiment words and negative situations.~\newcite{joshi2015harnessing} use multiple features comprising lexical, pragmatics, implicit and explicit context incongruity. In the explicit case, they include relevant features to detect thwarted sentimental expectations in the sentence. For implicit incongruity, they generalize~\newcite{riloff2013sarcasm}'s work in identifying verb-noun phrases containing contrast in both polarities.  

\paragraph{Context-based: }
Usage of contextual sarcasm has increased in the recent past, especially in online platforms. Texts found in microblogs, discussion forums, social media, etc., are plagued by grammatical inaccuracies and contain information which is highly temporal and contextual. In such scenarios, mining linguistic information becomes relatively inefficient and need arises for additional clues~\cite{carvalho2009clues}.~\newcite{wallace2014humans} demonstrate this need by showing how traditional classifiers fail in instances where humans require additional context. They also indicate the importance of speaker and/or topical information associated to a text to gather such context.~\newcite{poria2016deeper} use additional information by sentiment, emotional and personality representations of the input text. Previous works have mainly used historical posts of users to understand sarcastic tendencies~\cite{rajadesingan2015sarcasm,zhang2016tweet}.~\newcite{khattri2015your} try to find users' sentiments towards entities in their histories to find contrasting evidence.~\newcite{wallace2015sparse}  utilize sentiments and noun phrases used within a forum to gather context typical to that forum. Such forum based modeling simulates user-communities. Our work follows similar motivation where we explore context provided by user profiling and the topical knowledge embedded in the discourse of comments in discussion-forums (subreddits~\footnote{\url{https://www.reddit.com/reddits/}}).~\newcite{amir2016modelling} perform user modeling by learning embeddings that capture homophily. This work is closest to our approach given the fact that we too learn user embeddings to acquire context. However, we take a different approach that involve stylometric and personality description of the users. Empirical evidence shows that these proposed features are better than previous user modeling approaches. Moreover, we learn discourse features which has not been explored before in the context of this task.

\section{Method} \label{sec:method} 
\subsection{Task Definition}

The task involves detection of sarcasm for comments made in online discussion forums, i.e., Reddit. Let us denote the set $U = \{u_1, ..., u_{N_u} \}$ for $N_u$-users, where each user participates across a subset of $N_t$-discussion forums (subreddits). For a comment $C_{ij}$ made by the $i^{th}$ user $u_i$ in the $j^{th}$ discussion forum $t_j$, the objective is to predict whether the comment posted is sarcastic or not. 

\subsection{Summary of the Proposed Approach} \label{sec:approach}

Given the comment $C_{ij}$ to be classified, CASCADE leverages \textit{content} and \textit{context}-based information from the comment. For content-based modeling of $C_{ij}$, a CNN is used to generate the representation vector $\vec{\bm{c}}_{i,j}$ for a comment. CNNs generate abstract representations of text by extracting location-invariant local patterns. This vector $\vec{\bm{c}}_{i,j}$ captures both syntactic and semantic information useful for the task at hand. For contextual modeling, CASCADE first learns user embeddings and discourse features of all users and discussion forums, respectively (Section~\ref{sec:context}).  Following this phase, CASCADE then retrieves the learnt user embedding $\vec{\bm{u}}_{i}$ of user $u_i$ and discourse feature vector $\vec{\bm{t}}_{j}$ of forum $t_j$. Finally, all three vectors $\vec{\bm{c}}_{i,j}$, $\vec{\bm{u}}_{i}$, and $\vec{\bm{t}}_{j}$ are concatenated and used for the classification (Section~\ref{sec:prediction}). One might argue that instead of using one CNN, we could use multiple CNN (explained in~\cite{majumder2017deep}) to get better text representations whenever a comment contains multiple sentences. However that is out of the scope of this work. Here, we aim to show the effectiveness of user specific analysis and context-based features extracted from the discourse. Also the use of a single CNN for text representation helps to consistently compare with the state of the art.

\subsection{Learning Contextual Features} \label{sec:context}

We now detail the procedures to generate the contextual features, i.e., user embeddings and discourse features. The user embeddings try to capture users' traits that correlate to their sarcastic tendencies. These embeddings are created considering the accumulated historical posts of each user (Section~\ref{sec:user}). Contextual information are also extracted from the discourse of comments within each discussion forum. These extracted features are named as discourse features (Section~\ref{sec:discourse}). The aim of learning these contextual features is to acquire discriminative information crucial for sarcasm detection.

\subsection{User Embeddings} \label{sec:user}
To generate user embeddings,  we model their stylometric and personality features and then fuse them using CCA to create a single representation. Below we explain the generation of user embedding $\vec{\bm{u}}_{i}$, for the $i^{th}$ user $u_i$. Figure~\ref{fig:cca} also summarizes the overall architecture for this user profiling.

\subsubsection{Stylometric features} \label{sec:stylometric}
People possess their own idiolect and authorship styles, which is reflected in their writing. These styles are generally affected by attributes such as gender, diction, syntactic influences, etc.~\cite{cheng2011author,stamatatos2009survey} and present behavioral patterns which aid sarcasm detection~\cite{rajadesingan2015sarcasm}.  

We use this motivation to learn stylometric features of the users by consolidating their online comments into documents. We first gather all the comments by a user and create a document by appending them using a special delimiter \verb|<END>|. An unsupervised representation learning method \textit{ParagraphVector}~\cite{le2014distributed} is then applied on this document. This method generates a fixed-sized vector for each user by performing the auxiliary task of predicting the words within the documents. The choice of \textit{ParagraphVector} is governed by multiple reasons. Apart from its ability to effectively encode a user's writing style, it has the advantage of applying to variable lengths of text. 
\textit{ParagraphVector} also has been shown to perform well for sentiment classification tasks. The existence of synergy between sentiment and sarcastic orientation of a sentence also promotes the use of this method.

We now describe the functioning of this method. Every user-document and all words within them are first mapped to unique vectors such that each vector is represented by a column in matrix $D \in \mathbb{R}^{d_{s} \times N_u}$ and $W_s \in \mathbb{R}^{d_{s} \times |V|}$, respectively. Here, $d_s$ is the embedding size and $|V|$ represents the size of the vocabulary. \textit{Continuous-bag-of-words} approach~\cite{mikolov2013distributed} is then performed where a target word is predicted given the word vectors from its context-window. The key idea here is to use the document vector of the associated document as part of the context words. More formally, given a user-document $d_i$ for user $u_i$ comprising a sequence of $n_i$-words $\bm{w}_1, \bm{w}_2, ..., \bm{w}_{n_i}$, we calculate the average log probability of predicting each word within a sliding context window of size $k_s$. This average log probability is:
\begin{equation}
\frac{1}{n_i} \sum_{t=k_s}^{n_i - k_s}\text{log} \ p(w_t | d_{i}, w_{t-k_s},..., w_{t+k_s})
\end{equation}

To predict a word within a window, we take the average of all the neighboring context word vectors along with the document vector $\vec{\bm{d}}_{i}$ and use a neural network with softmax prediction: 
\begin{equation}
p(w_t | d_{i}, w_{t-k_s},..., w_{t+k_s}) = \frac{e^{\vec{\bm{y}}_{w_t}}}{\sum_ie^{\vec{\bm{y}}_i}}
\end{equation}

Here, $\vec{\bm{y}} = [y_1, ..., y_{|V|}]$ is the output of the neural network, i.e.,
\begin{equation}
\vec{\bm{y}} = \text{U}_dh(\vec{\bm{d}}_{i},\vec{\bm{w}}_{t-k_s},..., \vec{\bm{w}}_{t+k_s};D,W_s) + \vec{\bm{b}}_d
\end{equation}
$\vec{\bm{b}}_d \in \mathbb{R}^{|V|}, U_d \in \mathbb{R}^{|V| \times d_s}$ are parameters and $h(\cdot)$ represents the average of vectors $\vec{\bm{d}}_{i},\vec{\bm{w}}_{t-k_s},..., \vec{\bm{w}}_{t+k_s}$ taken from $D \ \text{and} \ W_s$. Hierarchical softmax is used for faster training~\cite{morin2005hierarchical}. Finally, after training, $D$ learns the users' document vectors which represent their stylometric features.


\subsubsection{Personality features} \label{sec:personality}

Discovering personality from text has numerous NLP applications such as product recognition, mental health diagnosis, etc.~\cite{majumder2017deep}. Described as a combination of multiple characteristics, personality detection helps in identifying behavior, thought patterns of an individual. To model the dependencies of users' personality with their sarcastic nature, we include personality features in the user embeddings.
Previously,~\newcite{poria2016deeper} also utilize personality features in sentences. However, we take a different and more-involved approach of extracting the personality features of a user instead.

For user $u_i$, we iterate over all the $v_i$-comments \{$S_{u_i}^1, ..., S_{u_i}^{v_i}$\} written by them. For each $S_{u_i}^j$, we provide the comment as an input to a pre-trained Convolutional Neural Network (CNN) which has been trained on a multi-label personality detection task. Specifically, the CNN is pre-trained on a benchmark corpus developed by~\newcite{matthews1999personality} which contains $2,400$ essays and is labeled with the Big-Five personality traits, i.e., Openness, Conscientiousness, Extraversion, Agreeableness, and Neuroticism (OCEAN). After the training, this CNN model is used to infer the personality traits present in each comment. This is done by extracting the activations of the CNN's last hidden layer vector which we call as the personality vector {${\vec{\bm{p}}_{u_i}}^j$. The expectation over the personality vectors for all $v_i$-comments made by the user is then defined as the overall personality feature vector $\bm{\vec{p}}_{i}$ of user $u_i$:
\begin{equation}
\bm{\vec{p}}_{i} \ = \ \mathbb{E}_{j\in [v_i]} [ \vec{\bm{p}}_{u_i}^j ] \ = \ \frac{1}{v_i}\sum_{j=1}^{v_i} \vec{\bm{p}}_{u_i}^j
\end{equation}
\paragraph{CNN:}
Here, we describe the CNN that generates the personality vectors. Given a user's comment, which is a text $S = [w_1, ..., w_n]$ composed of $n$ words, each word $w_i$ is represented as a word embedding $\vec{\bm{w}}_i \in \mathbb{R}^{d_{em}}$  using the pre-trained FastText embeddings~\cite{bojanowski2016enriching}. A single-layered CNN is then modeled on this input sequence $S$~\cite{kim2014convolutional}. First, a convolutional layer is applied having three filters $F_{[1,2,3]} \in \mathbb{R}^{d_{em} \times h_{[1,2,3]}}$ of heights $h_{[1,2,3]}$, respectively. For each $k \in \{1,2,3 \}$, filter $F_k$ slides across $S$ and extracts $h_k$-gram features at each instance.  This creates a feature map vector $\vec{\bm{m}}_k$ of size $\mathbb{R}^{\mid S \mid - h_k + 1}$, whose each entry $m_{k,j}$ is obtained as:
\begin{equation}
m_{k,j} = \alpha ( \ {F_k \cdot S_{[j:j+h_k-1]} + b_k} \ )
\end{equation}
here, $b_k \in \mathbb{R}$ is the bias and $\alpha(\cdot)$ is a non-linear activation function.

$M$ feature maps are created from each filter $F_k$ giving a total of $3M$ feature maps as output. Following this, a max-pooling operation is performed across the length of each feature map. Thus, for all $M$ feature maps computed from $F_k$, output $\vec{\bm{o_k}}$ is calculated as, $ \vec{\bm{o_k}} = [ \text{ \footnotesize \textit{max}}(\vec{\bm{m}}_1^1), ..., \text{ \footnotesize \textit{max}}(\vec{\bm{m}}_1^{ \text{\small M}}) \ ] $. Overall the max-pooling output is calculated by concatenation of each $\vec{\bm{o_k}}$ to get $\vec{\bm{o}} = [\vec{\bm{o_1}} \oplus \vec{\bm{o_2}} \oplus \vec{\bm{o_3}}] \in \mathbb{R}^{3M}$ , where $\oplus$ represents concatenation. Finally, $\vec{\bm{o}}$ is projected onto a dense layer with $d_p$ neurons followed by the final sigmoid-prediction layer with $5$ classes denoting the five personality traits~\cite{matthews2003personality}. We use sigmoid instead of softmax to facilitate multi-label classification. This is calculated as,
\begin{gather}
\bm{\hat{y}} = \sigma( \ W_2\vec{\bm{q}} + \vec{\bm{b}}_2 \ ) \quad \text{, where} \quad \vec{\bm{q}}  = \alpha( \ W_1\vec{\bm{o}} + \vec{\bm{b}}_1 \ ) 
\end{gather}
$W_1\in \mathbb{R}^{d_p \times 3M}, W_2 \in \mathbb{R}^{5 \times d_p}, \vec{\bm{b}}_1 \in \mathbb{R}^{d_p}$ and $\vec{\bm{b}}_2 \in \mathbb{R}^{5}$ are parameters and $\alpha(.)$ represents non-linear activation.

\begin{figure}[t] 
	\centering 
	\includegraphics[width = 0.95\textwidth]{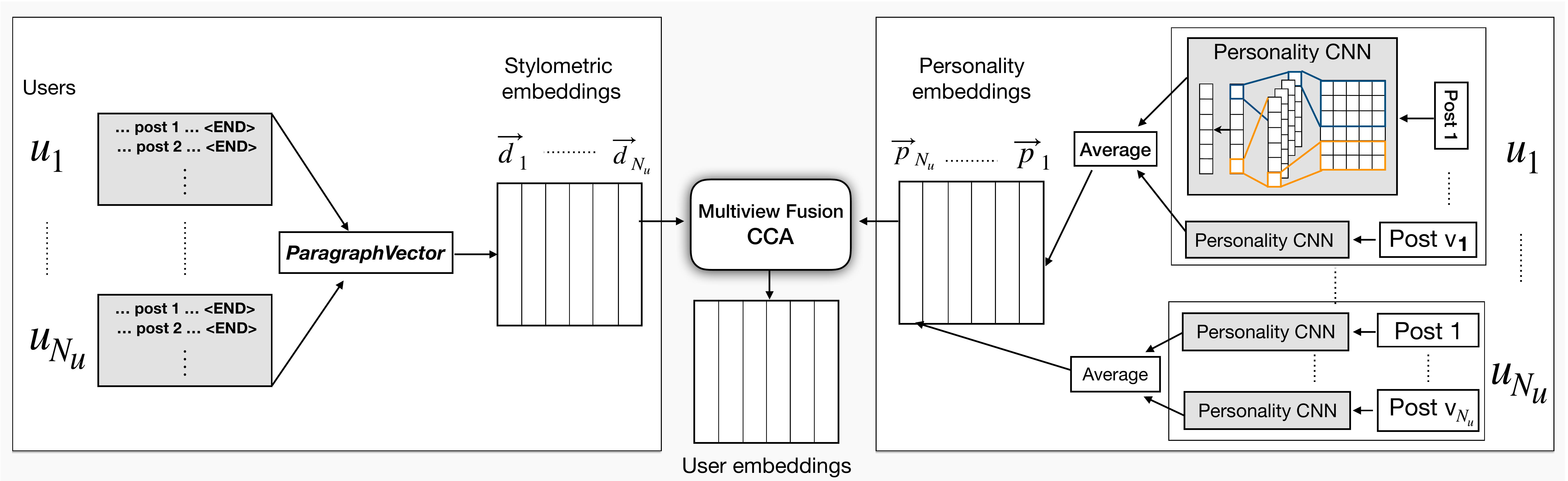}   
	\caption[]{\footnotesize The figure describes the process of user profiling. Stylometric and Personality embeddings are generated and then fused in a multi-view setting using CCA to get the user embeddings.}
	\label{fig:cca}
    \vspace{-0.4cm}
 \end{figure}

\subsubsection{Fusion} \label{sec:fusion}
We take a multi-view learning approach to combine both stylometric and personality features into a comprehensive embedding for each user. We use \textit{Canonical Correlation Analysis} (CCA)~\cite{hotelling1936relations} to perform this fusion. CCA captures maximal information between two views and creates a combined representation~\cite{hardoon2004canonical,benton2016learning}. In the event of having more than two views, fusion can be performed using an extension of CCA called \textit{Generalized} CCA (see Supplementary).

\paragraph{Canonical Correlation Analysis:}
Let us consider the learnt stylometric embedding matrix $D \in \mathbb{R}^{d_s \times N_u}$ and personality embedding matrix  $P \in \mathbb{R}^{d_p \times N_u}$ containing the respective embedding vectors of user $u_i$ in their $i^{th}$ columns. The matrices are then mean-centered and standardized across all user columns. We call these new matrices as 
$X_1$ and $X_2$, respectively. Let the correlation matrix for $X_1$ be  $R_{11} = X_1{X_1}^T \in \mathbb{R}^{d_s \times d_s}$, for $X_2$  be $R_{22} = X_2{X_2}^T \in \mathbb{R}^{d_p \times d_p}$ and  the cross-correlation matrix between them be $R_{12} = X_1{X_2}^T \in \mathbb{R}^{d_s \times d_p}$. For each user $u_i$, the objective of CCA is to find the linear projections of both embedding vectors that have a maximum correlation. We create $K$ such projections, i.e., $K$-canonical variate pairs such that each pair of projection is orthogonal with respect to the previous pairs. This is done by constructing:
\begin{equation}
W = X_1^TA_1 \ \text{and}  \ Z = X_2^TA_2
\end{equation}
where, $A_1 \in \mathbb{R}^{d_s \times K}$, $A_2 \in \mathbb{R}^{d_p \times K}$ and $W^TW = Z^TZ = I$. To maximize correlation between $W$ and $Z$, optimal $A_1$ and $A_2$ are calculated by performing singular value decomposition as:
\begin{equation}
 R_{11}^{-\frac{1}{2}}R_{12}R_{22}^{-\frac{1}{2}} = A\Lambda B^\top  \quad \text{, where} \quad A_1 = R_{11}^{-\frac{1}{2}} A \textnormal{ and } A_2 = R_{22}^{-\frac{1}{2}}B
\end{equation}
It can be seen that,
\begin{gather}
W^TW = {A_1}^TR_{11}{A_1} = A^TA =  I \quad \text{and} \quad Z^TZ = {A_2}^TR_{22}{A_2} = B^TB =  I \\
\text{also,} \quad W^TZ = Z^TW = \Lambda
\end{gather}
Once optimal $A_1$ and $A_2$ are calculated, overall user embedding $\vec{\bm{u}}_{i} \in \mathbb{R}^{K}$ of user $u_i$ is generated by fusion of $\bm{\vec{d}}_{i}$ and $\bm{\vec{p}}_{i}$ as: 
\begin{equation}
\vec{\bm{u}}_{i} = {(\bm{\vec{d}}_{i})}^TA_1 + {(\bm{\vec{p}}_{i})}^TA_2 
\end{equation}

\vspace{-0.3cm}
\subsection{Discourse Features} \label{sec:discourse}

Similar to how a user influences the degree of sarcasm in a comment, we assume that the discourse of  comments belonging to a certain discussion forum contain contextual information relevant to the sarcasm classification. They embed topical information that selectively incur bias towards degree of sarcasm in the comments of a discussion. For example, comments on political leaders or sports matches are generally more susceptible to sarcasm than natural disasters. Contextual information extracted from the discourse of a discussion can also provide background knowledge or cues about the topic of that discussion.

To extract the discourse features, we take a similar approach of document modeling performed for stylometric features (Section~\ref{sec:stylometric}). For all $N_t$-discussion forums, we compose each forum's document by appending the comments within them. As before, \textit{ParagraphVector} is employed to generate discourse representations for each document. We denote the learnt feature vector of $j^{th}$ forum $t_j$ as $\vec{\bm{t}}_{j} \in \mathbb{R}^{d_t}$.

\begin{figure}[t] 
	\centering 
	\includegraphics[width = 0.79\textwidth]{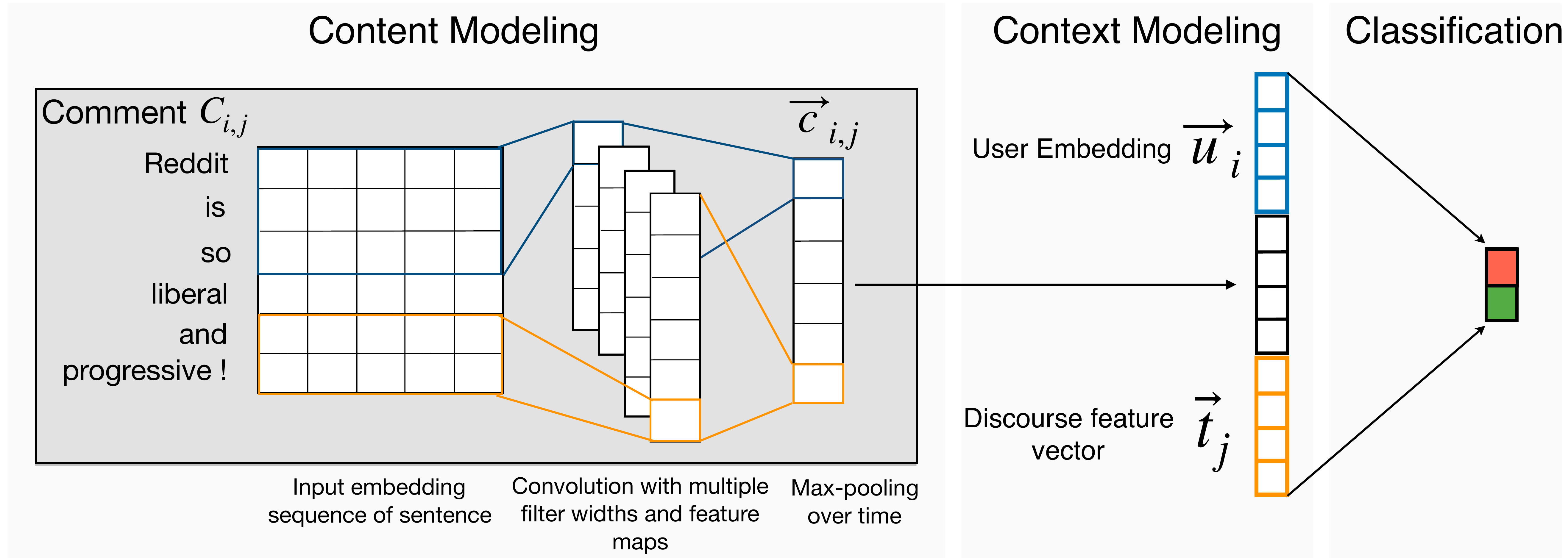}   
	\caption[]{\footnotesize This figure presents the overall hybrid network CASCADE. For the comment $C_{i,j}$, its content-based sentential representation $\vec{\bm{c}}_{i,j}$ is extracted using a CNN and appended with context vectors   $\vec{\bm{u}}_{i}$ and $\vec{\bm{t}}_{j}$.}
	\label{fig:overall}
    \vspace{-0.3cm}
\end{figure}

\subsection{Final Prediction} \label{sec:prediction}

Following the extraction of text representation $\vec{\bm{c}}_{i,j}$ for comment $C_{i,j}$ and retrieval of user embedding $\vec{\bm{u}}_{i}$ for author $u_i$ and discourse feature vector $\vec{\bm{t}}_{j}$ for discussion forum $t_j$, we concatenate all three vectors to form the unified text representation $\hat{\bm{c}}_{i,j} = [\vec{\bm{c}}_{i,j} \oplus \vec{\bm{u}}_{i} \oplus \vec{\bm{t}}_{j}]$.  Here, $\oplus$ refers to concatenation. The CNN used for extraction of $\vec{\bm{c}}_{i,j}$ has the same design as the CNN we used to extract personality features described in Section~\ref{sec:personality}. Finally, $\hat{\bm{c}}_{i,j}$ is projected to the output layer having two neurons with a softmax activation. This gives a softmax-probability over whether a comment is sarcastic or not. This probability estimate is then used to calculate the categorical cross-entropy which is used as the loss function:
\begin{equation}
Loss = \frac{-1}{N}{\sum_{i=1}^{N}} {\sum_{j=1}^{2}}{\bm{y}_{i,j} \ log_2 (\hat{\bm{y}}_{i,j} )} \quad \text{, where} \quad \hat{\bm{y}} = softmax( W_o\hat{\bm{c}}_{i,j} + \vec{\bm{b}}_o )
\end{equation}
Here, $N$ is the number of comments in the training set, $y_{i}$ is the one-hot vector ground truth of the $i^{th}$ comment and $\hat{\bm{y}}_{i,j} $ is its predicted probability of belonging to class $j$.

\section{Experimental Results} \label{sec:experiments}
\subsection{Dataset}\label{sec:dataset}

We perform our experiments on a large-scale self-annotated corpus for sarcasm, \verb|SARC|\footnote{http://nlp.cs.princeton.edu/SARC/}~\cite{khodak2017large}. This dataset contains more than a million examples of sarcastic/non-sarcastic statements made in the social media site Reddit. Reddit comprises of topic-specific discussion forums, also known as subreddits, each titled by a post. In each forum, users communicate either by commenting to the titled post or other's comments, resulting in a tree-like conversation structure. This structure can be unraveled to a linear format, thus creating a discourse of the comments by keeping the topological constraints intact. Each comment is accompanied with its author details and parent comments (if any) which is subsequently used for our contextual processing. It is important to note that almost all comments in the SARC dataset are composed of a single sentence. We consider three variants of the \verb|SARC| dataset in our experiments. 
\begin{itemize}[leftmargin=*]
\itemsep0em 
\item \textbf{Main balanced:} This is the primary dataset which contains a balanced distribution of both sarcastic and non-sarcastic comments. The dataset contains comments from $1246058$ users ($118940$ in training and $56118$ in testing set) distributed across $6534$ forums ($3868$ in training and $2666$ in testing set).
\item \textbf{Main imbalanced:} To emulate real-world scenarios where the sarcastic comments are typically lesser than non-sarcastic ones, we use an imbalanced version of the Main dataset. Specifically, we maintain a $20:80$ ratio (approx.) between the sarcastic and non-sarcastic comments in both training/testing sets.
\item \textbf{Pol:} To further test the effectiveness of our user embeddings, we perform experiments on a subset of Main, comprising of forums associated with the topic of politics. Table~\ref{table:SARCstats} provides the comment distribution of all the dataset variants mentioned. 
\end{itemize}
\begin{table}[h!]
	\small
	\begin{center}
		\begin{tabular}{c|c|c|c|c|c|c|c|c|c}
			\Xhline{3\arrayrulewidth}
			\multicolumn{2}{c|}{}&\multicolumn{4}{c}{Training set}&\multicolumn{4}{|c}{Testing set}\\ \cline{3-10}
            \multicolumn{2}{c|}{}&\multicolumn{2}{c|}{\multirow{2}{*}{no. of comments}}&\multicolumn{2}{c|}{avg. no. of words }&\multicolumn{2}{c|}{\multirow{2}{*}{no. of comments}}&\multicolumn{2}{c}{avg. no. of words}\\ 
            
            \multicolumn{2}{c|}{}&\multicolumn{2}{c|}{}&\multicolumn{2}{c|}{per comment}&\multicolumn{2}{c|}{}&\multicolumn{2}{c}{per comment}\\\cline{3-10}
		   	\multicolumn{2}{c|}{}&\textit{non-sarc}&\textit{sarc}&\textit{non-sarc}&\textit{sarc}&\textit{non-sarc}&\textit{sarc}&\textit{non-sarc}&\textit{sarc}\\ 
            \hline \hline
            \multirow{2}{*}{Main}&balanced&77351&77351&55.13&55.08&32333&32333&55.55&55.01\\
            &imbalanced&77351&25784&55.13&55.21&32333&10778&55.55&55.48\\ \cline{2-2}
            Pol&balanced&6834&6834&64.74&62.36&1703&1703&62.99&62.14\\
            \hline
            \multicolumn{10}{c}{\footnotesize{$^*$non-sarc: non-sarcastic, sarc: sarcastic}}
		\end{tabular}
	\end{center}
	\vspace{-0.4cm}
	\caption {Details of comments in the SARC datasets.}
	\label{table:SARCstats}
\end{table} 

The choice of using \verb|SARC| for our experiments comes with multiple reasons. First, this corpus is the first of its kind that was purposely developed to investigate the necessity of contextual information in sarcasm classification. This characteristic aligns well with the main goal of this paper. Second, the large size of the corpus allows for statistically-relevant analyses. Third, the dataset annotations contain a small false-positive rate for sarcastic labels thus providing reliable annotations. Also, its self-annotation scheme rules out the annotation errors induced by third-party annotators. Finally, the corpus structure provides meta-data (e.g., user information) for its comments, which is useful for contextual modeling.

\subsection{Training details} \label{sec:training}
We hold out 10\% of the training data for validation. Hyper-parameter tuning is performed using this validation set through RandomSearch~\cite{bergstra2012random}. To optimize the parameters, Adam optimizer~\cite{kingma2014adam} is used, starting with an initial learning rate of $1e^{-4}$. The learnable parameters in the network consists of $\theta = \{ U_d, D, W_{[1,2,o,s]}, F_{[1,2,3]}, \vec{\textbf{b}}_{[1,2,o,d]}, b_{[1,2,3]} \}$. Training termination is decided using early stopping technique with a patience of $12$. For the batched-modeling of comments in CNNs, each comment is either restricted or padded to $100$ words for uniformity. The optimal hyper-parameters are found to be $\{d_s, d_p, d_t, K\} = 100$, $d_{em} = 300$, $k_s = 2$, $M = 128$, and $\alpha = ReLU$ (Implementation details are provided in the supplementary).



\subsection{Baseline Models} \label{sec:baselines}
Here we describe the state-of-the-art methods and baselines that we compare CASCADE with. 
\begin{itemize}[leftmargin=*,noitemsep]
	\item \textbf{Bag-of-Words:} This model uses a comment's word-counts as features in a vector. The size of the vector is the vocabulary size of the training dataset.
	\item \textbf{CNN:} We compare our model with this individual CNN version. This CNN is capable of modeling only the \textit{content} of a comment. The architecture is similar to the CNN used in CASCADE (see Section~\ref{sec:approach}).
    \item \textbf{CNN-SVM:} This model proposed by~\newcite{poria2016deeper} consists of a CNN for content modeling and other pre-trained CNNs for extracting sentiment, emotion and personality features from the given comment. All the features are concatenated and fed into an SVM for classification. 
     \item \textbf{CUE-CNN:} This method proposed by~\newcite{amir2016modelling} also models user embeddings with a method akin to \textit{ParagraphVector}. Their embeddings are then combined with a CNN thus forming the CUE-CNN model. We compare with this model to analyze the efficiency of our embeddings as opposed to theirs. Released software\footnote{\protect\url{https://github.com/samiroid/CUE-CNN}} is used to produce results on the SARC dataset.
\end{itemize}

\subsection{Results} \label{sec:results}
\begin{table}[t!]
	\small
	\newcommand\Y{\hphantom{$^1$}}
	\newcommand\X{\%\Y}
	\begin{center}
		\begin{tabular}{lc|c|c|c|c|c|c}
			\Xhline{3\arrayrulewidth}
			\multicolumn{2}{l|}{\multirow{3}{*}{Models}}&\multicolumn{4}{|c}{Main}&\multicolumn{2}{|c}{Pol} \\ \cline{3-8}
		   	&&\multicolumn{2}{|c}{balanced}&\multicolumn{2}{|c|}{imbalanced}&\multicolumn{2}{|c}{}\\ \cline{3-8}
            &&\scriptsize{\textit{Accuracy}}&\scriptsize{\textit{F1}}&\scriptsize{\textit{Accuracy}}&\scriptsize{\textit{F1}}&\scriptsize{\textit{Accuracy}}& \scriptsize{\textit{F1}}\\ \hline \hline
            Bag-of-Words&&0.63&0.64&0.68&0.76&0.59&0.60\\
            CNN&&0.65&0.66&0.69&0.78&0.62&0.63\\
            CNN-SVM~\cite{poria2016deeper}&&0.68&0.68&0.69&0.79&0.65&0.67\\ 
            CUE-CNN~\cite{amir2016modelling}&&0.70&0.69&0.73&0.81&0.69&0.70\\ \hline
            CASCADE (no personality features)&&0.68&0.66&0.71&0.80&0.68&0.70\\
            
      CASCADE&&\textbf{0.77}$^\dagger$&\textbf{0.77}$^\dagger$&\textbf{0.79}$^\dagger$&\textbf{0.86}$^\dagger$&\textbf{0.74}$^\dagger$&\textbf{0.75}$^\dagger$\\ \hline \hline
            $\Delta_{SOTA}$&& \textcolor{mygreen}{$\uparrow 7\%$} & \textcolor{mygreen}{$\uparrow 8\%$} & \textcolor{mygreen}{$\uparrow 6\%$} & \textcolor{mygreen}{$\uparrow 5\%$} & \textcolor{mygreen}{$\uparrow 5\%$} & \textcolor{mygreen}{$\uparrow 5\%$} \\ \Xhline{3\arrayrulewidth} 
            \multicolumn{8}{l}{\scriptsize{$\dagger$}:significantly better than CUE-CNN~\cite{amir2016modelling}.}
		\end{tabular}
	\end{center} 
	
	\vspace{-0.4cm}
	\caption {Comparison of CASCADE with state-of-the-art networks and baselines on multiple versions of the SARC dataset. We assert significance when $p < 0.05$ under paired-t test. Results comprise of $10$ runs with different initializations. The bottom row shows the absolute difference with respect to the CUE-CNN system.}
	\label{table:results}
\end{table} 

Table~\ref{table:results} presents the performance results on the SARC datasets. CASCADE manages to achieve major improvement across all datasets with statistical significance. The lowest performance is obtained by the Bag-of-words approach whereas all neural architectures outperform it. Amongst the neural networks, the CNN baseline receives the least performance. CASCADE comfortably beats the state-of-the-art neural models CNN-SVM and CUE-CNN. Its improved performance on the Main imbalanced dataset also reflects its robustness towards class imbalance and establishes it as a real-world deployable network.

We further compare our proposed user-profiling method with that of CUE-CNN, with absolute differences shown in the bottom row of Table~\ref{table:results}.  Since CUE-CNN generates its user embeddings using a method similar to the \textit{ParagraphVector}, we test the importance of personality features being included in our user profiling. As seen in the table, CASCADE without personality features drops in performance to a range similar to CUE-CNN. This suggests that the combination of stylometric and personality features are indeed crucial for the improved performance of CASCADE.

\begin{figure}[b!]
\begin{floatrow}
\capbtabbox{%

		\small
 		\setlength\tabcolsep{3pt}
		\begin{tabular}{l|c|c|c|c|c|c|c|c|r}
			\Xhline{3\arrayrulewidth}
			\multicolumn{4}{c|}{CASCADE}&\multicolumn{4}{|c}{Main}&\multicolumn{2}{|c}{Pol} \\ \cline{1-9}
		   	&\multicolumn{2}{c|}{user}&\footnotesize{dis-}&\multicolumn{2}{|c}{\footnotesize{balanced}}&\multicolumn{2}{|c|}{\footnotesize{imbalanced}}&\multicolumn{2}{|c}{}\\ \cline{2-3} \cline{5-10}
            &\scriptsize{cca}&\scriptsize{concat.}&\footnotesize{course}&\scriptsize{\textit{Acc.}}&\scriptsize{\textit{F1}}&\scriptsize{\textit{Acc.}}&\scriptsize{\textit{F1}}&\scriptsize{\textit{Acc.}}& \scriptsize{\textit{F1}}\\ \hline \hline
            1.&-&-&-&0.65&0.66&0.69&0.78&0.62&0.63\\
            2.&-&-& \cmark &0.66&0.66&0.68&0.78&0.63&0.66\\ 
            3.&-&\cmark & - &0.66&0.66&0.69&0.79&0.62&0.61\\
            4.&-&\cmark & \cmark &0.65&0.67&0.71&0.85&0.63&0.66\\
            5.&\cmark&-&-&0.77&0.76&\textbf{0.80}&\textbf{0.86}&0.70&0.70\\
            6.&\cmark & -& \cmark &\textbf{0.78}&\textbf{0.77}&0.79&\textbf{0.86}&\textbf{0.74}&\textbf{0.75}\\ \hline
            
		\end{tabular}
  		\vspace{0.5cm}
  
}{%
  \caption{\footnotesize{Comparison with variants of the proposed CASCADE network. All combinations use content-based CNN.}}%
  \label{table:variationresults}
}

\ffigbox[\FBwidth]{%
	\includegraphics[width = 0.45\textwidth]{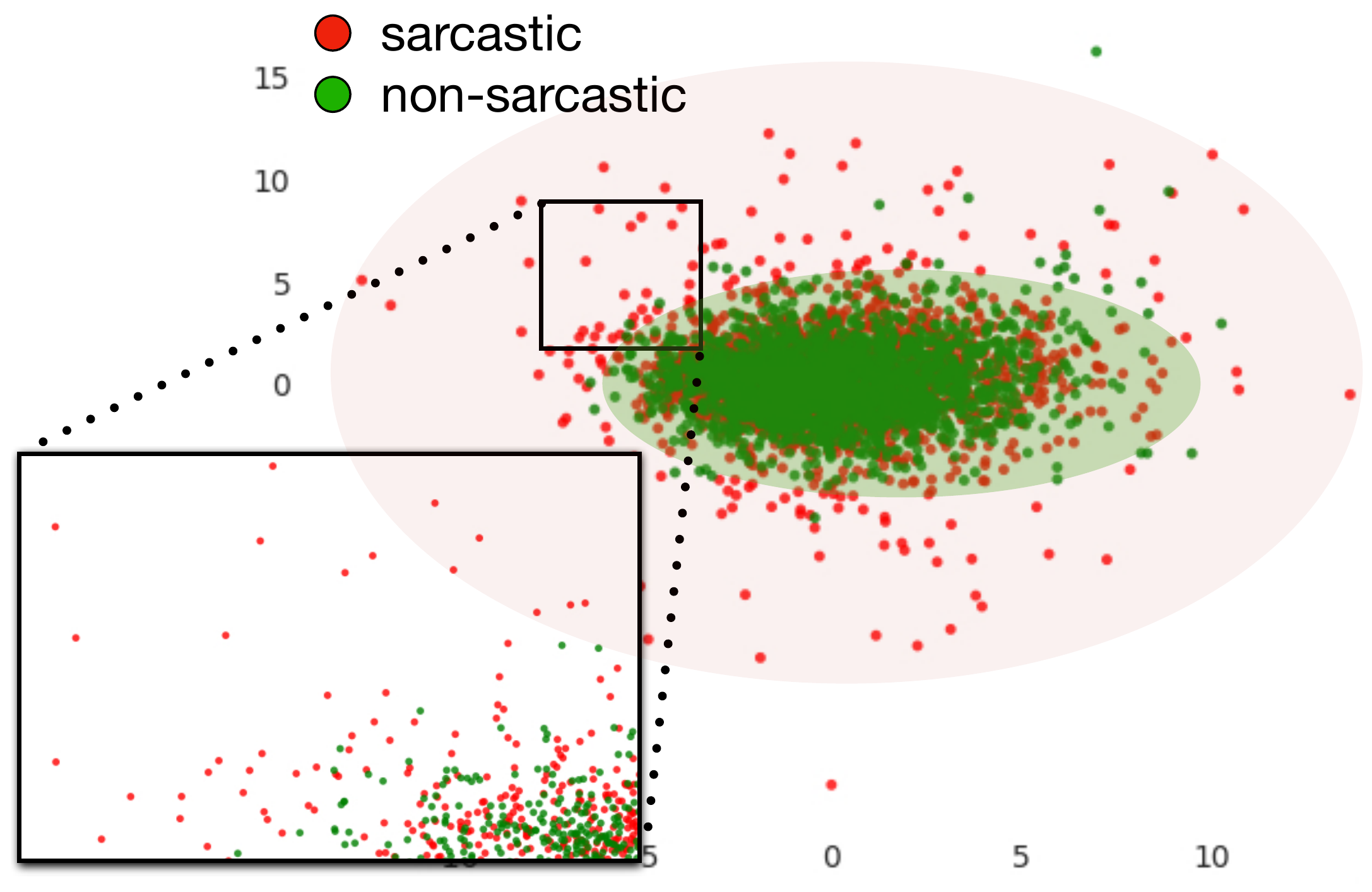}
    
  }{%
    \caption{\footnotesize{2D-Scatterplot of the user embeddings visualized using t-SNE~\cite{maaten2008visualizing}.}}%
    \label{fig:visual1}
  }
\end{floatrow}
\end{figure}

\subsection{Ablation Study} \label{sec:ablation}

We experiment on multiple variants of CASCADE so as to analyze the importance of the various features present in its architecture. Table~\ref{table:variationresults} provides the results of all the combinations. First, we test performance for the \textit{content}-based CNN only (row~1). This setting provides the worst relative performance with almost $10\%$ lesser accuracy than optimal. Next, we include contextual features to this network. Here, the effect of discourse features is primarily seen in the Pol dataset getting an increase of $3\%$ in F1 (row~2). A major boost in performance is observed ($8-12\%$ accuracy and F1) when user embeddings are introduced (row~5). Visualization of the user embedding cluster (Section~\ref{sec:visualization}) provides insights for this positive trend. Overall, CASCADE consisting of CNN with user embeddings and contextual discourse features provide the best performance in all three datasets (row~6).

We challenge the use of CCA for the generation of user embeddings and thus replace it with simple concatenation. This however causes a significant drop in performance (row~3). Improvement is not observed even when discourse features are used with these concatenated user embeddings (row~4). We assume the increase in parameters caused by concatenation for this performance degradation. CCA on the other hand creates succinct representations with maximal information, giving better results. 

\subsection{User Embedding Analysis} \label{sec:visualization}

We investigate the learnt user embeddings in more detail. In particular, we plot random samples of users on a 2D-plane using t-SNE~\cite{maaten2008visualizing}. The users who have greater sarcastic comments (atleast $2$ more than the other type) are termed as sarcastic users (colored red). Conversely, the users having lesser sarcastic comments are called non-sarcastic users (colored green). Equal number of users from both the categories are plotted. We aim to analyze the reason behind the performance boost provided by the user embeddings as shown in Table~\ref{table:variationresults}. We see in Figure~\ref{fig:visual1} that both the user types belong to similar distributions. However, the sarcastic users have a greater spread than the non-sarcastic ones (red belt around the green region). This is also evident from the variances of the distributions where the sarcastic distribution comprises of $10.92$ variance as opposed to $5.20$ variance of the non-sarcastic distribution. We can infer from this observation that the user embeddings belonging to this non-overlapping red-region provide discriminative information regarding the sarcastic tendencies of their users.

\subsection{Case Studies} \label{sec:casestudies}

Results demonstrate that discourse features provide an improvement over baselines, especially on the Pol dataset. This signifies the greater role of the contextual cues for classifying comments in this dataset over the other dataset variants used in our experiment.
Below, we present a couple of cases from the Pol dataset where our model correctly identifies the sarcasm which is evident only with the neighboring comments. The previous state-of-the-art CUE-CNN, however, misclassifies them.
\begin{itemize}[leftmargin=*,noitemsep]
    \item For the comment \textit{Whew, I feel much better now!}, its sarcasm is evident only when its previous comment is seen \textit{So all of the US presidents are terrorists for the last 5 years.}
	\item The comment \textit{The part where Obama signed it.} doesn't seem to be sarcastic until looked upon as a remark to its previous comment \textit{What part of this would be unconstitutional?}.
\end{itemize}
Such observations indicate the impact of discourse features. However, sometimes contextual cues from the previous comments are not enough and misclassifications are observed due to lack of necessary commonsense and background knowledge about the topic of discussion. There are also other cases where our model fails despite the presence of contextual information from the previous comments. During exploration, this is primarily observed for contextual comments which are very long. Thus, sequential discourse modeling using RNNs may be better suited for such cases. Also, in the case of user embeddings, misclassifications were common for users with lesser historical posts. In such scenarios, potential solutions would be to create user networks and derive information from similar users within the network.
These are some of the issues which we plan to address in future work.

\section{Conclusion} \label{sec:conclusion}
In this paper we introduce Contextual Sarcasm Detector called as CASCADE which leverages both content and contextual information for the classification. For contextual details, we perform user profiling along with discourse modeling from comments in discussion threads. When this information is used jointly with a CNN-based textual model, we obtain state-of-the-art performance on a large-scale Reddit corpus. Our results show that discourse features along with user embeddings play a crucial role in the performance of sarcasm detection. 

\bibliographystyle{acl}
\bibliography{coling2018}

\begin{thebibliography}{}

\bibitem[\protect\citename{Amir \bgroup et al.\egroup }2016]{amir2016modelling}
Silvio Amir, Byron~C Wallace, Hao Lyu, and Paula Carvalho M{\'a}rio~J Silva.
\newblock 2016.
\newblock Modelling context with user embeddings for sarcasm detection in
  social media.
\newblock {\em arXiv preprint arXiv:1607.00976}.

\bibitem[\protect\citename{Benton \bgroup et al.\egroup
  }2016]{benton2016learning}
Adrian Benton, Raman Arora, and Mark Dredze.
\newblock 2016.
\newblock Learning multiview embeddings of twitter users.
\newblock In {\em Proceedings of the 54th Annual Meeting of the Association for
  Computational Linguistics (Volume 2: Short Papers)}, volume~2, pages 14--19.

\bibitem[\protect\citename{Bergstra and Bengio}2012]{bergstra2012random}
James Bergstra and Yoshua Bengio.
\newblock 2012.
\newblock Random search for hyper-parameter optimization.
\newblock {\em Journal of Machine Learning Research}, 13(Feb):281--305.

\bibitem[\protect\citename{Bojanowski \bgroup et al.\egroup
  }2016]{bojanowski2016enriching}
Piotr Bojanowski, Edouard Grave, Armand Joulin, and Tomas Mikolov.
\newblock 2016.
\newblock Enriching word vectors with subword information.
\newblock {\em arXiv preprint arXiv:1607.04606}.

\bibitem[\protect\citename{Carvalho \bgroup et al.\egroup
  }2009]{carvalho2009clues}
Paula Carvalho, Lu{\'\i}s Sarmento, M{\'a}rio~J Silva, and Eug{\'e}nio
  De~Oliveira.
\newblock 2009.
\newblock Clues for detecting irony in user-generated contents: oh...!! it's so
  easy;-.
\newblock In {\em Proceedings of the 1st international CIKM workshop on
  Topic-sentiment analysis for mass opinion}, pages 53--56. ACM.

\bibitem[\protect\citename{Cheng \bgroup et al.\egroup }2011]{cheng2011author}
Na~Cheng, Rajarathnam Chandramouli, and KP~Subbalakshmi.
\newblock 2011.
\newblock Author gender identification from text.
\newblock {\em Digital Investigation}, 8(1):78--88.

\bibitem[\protect\citename{Davidov \bgroup et al.\egroup
  }2010]{davidov2010semi}
Dmitry Davidov, Oren Tsur, and Ari Rappoport.
\newblock 2010.
\newblock Semi-supervised recognition of sarcastic sentences in twitter and
  amazon.
\newblock In {\em Proceedings of the fourteenth conference on computational
  natural language learning}, pages 107--116. Association for Computational
  Linguistics.

\bibitem[\protect\citename{Gonz{\'a}lez-Ib{\'a}nez \bgroup et al.\egroup
  }2011]{gonzalez2011identifying}
Roberto Gonz{\'a}lez-Ib{\'a}nez, Smaranda Muresan, and Nina Wacholder.
\newblock 2011.
\newblock Identifying sarcasm in twitter: a closer look.
\newblock In {\em Proceedings of the 49th Annual Meeting of the Association for
  Computational Linguistics: Human Language Technologies: Short Papers-Volume
  2}, pages 581--586. Association for Computational Linguistics.

\bibitem[\protect\citename{Hardoon \bgroup et al.\egroup
  }2004]{hardoon2004canonical}
David~R Hardoon, Sandor Szedmak, and John Shawe-Taylor.
\newblock 2004.
\newblock Canonical correlation analysis: An overview with application to
  learning methods.
\newblock {\em Neural computation}, 16(12):2639--2664.

\bibitem[\protect\citename{Hotelling}1936]{hotelling1936relations}
Harold Hotelling.
\newblock 1936.
\newblock Relations between two sets of variates.
\newblock {\em Biometrika}, 28(3/4):321--377.

\bibitem[\protect\citename{Joshi \bgroup et al.\egroup
  }2015]{joshi2015harnessing}
Aditya Joshi, Vinita Sharma, and Pushpak Bhattacharyya.
\newblock 2015.
\newblock Harnessing context incongruity for sarcasm detection.
\newblock In {\em Proceedings of the 53rd Annual Meeting of the Association for
  Computational Linguistics and the 7th International Joint Conference on
  Natural Language Processing (Volume 2: Short Papers)}, volume~2, pages
  757--762.

\bibitem[\protect\citename{Joshi \bgroup et al.\egroup
  }2017]{joshi2017automatic}
Aditya Joshi, Pushpak Bhattacharyya, and Mark~J Carman.
\newblock 2017.
\newblock Automatic sarcasm detection: A survey.
\newblock {\em ACM Computing Surveys (CSUR)}, 50(5):73.

\bibitem[\protect\citename{Khattri \bgroup et al.\egroup
  }2015]{khattri2015your}
Anupam Khattri, Aditya Joshi, Pushpak Bhattacharyya, and Mark Carman.
\newblock 2015.
\newblock Your sentiment precedes you: Using an author’s historical tweets to
  predict sarcasm.
\newblock In {\em Proceedings of the 6th Workshop on Computational Approaches
  to Subjectivity, Sentiment and Social Media Analysis}, pages 25--30.

\bibitem[\protect\citename{Khodak \bgroup et al.\egroup }2017]{khodak2017large}
Mikhail Khodak, Nikunj Saunshi, and Kiran Vodrahalli.
\newblock 2017.
\newblock A large self-annotated corpus for sarcasm.
\newblock {\em arXiv preprint arXiv:1704.05579}.

\bibitem[\protect\citename{Kim}2014]{kim2014convolutional}
Yoon Kim.
\newblock 2014.
\newblock Convolutional neural networks for sentence classification.
\newblock In {\em Proceedings of the 2014 Conference on Empirical Methods in
  Natural Language Processing (EMNLP)}, pages 1746--1751.

\bibitem[\protect\citename{Kingma and Ba}2014]{kingma2014adam}
Diederik~P Kingma and Jimmy Ba.
\newblock 2014.
\newblock Adam: A method for stochastic optimization.
\newblock {\em arXiv preprint arXiv:1412.6980}.

\bibitem[\protect\citename{Kreuz and Caucci}2007]{kreuz2007lexical}
Roger~J Kreuz and Gina~M Caucci.
\newblock 2007.
\newblock Lexical influences on the perception of sarcasm.
\newblock In {\em Proceedings of the Workshop on computational approaches to
  Figurative Language}, pages 1--4. Association for Computational Linguistics.

\bibitem[\protect\citename{Le and Mikolov}2014]{le2014distributed}
Quoc Le and Tomas Mikolov.
\newblock 2014.
\newblock Distributed representations of sentences and documents.
\newblock In {\em Proceedings of the 31st International Conference on Machine
  Learning (ICML-14)}, pages 1188--1196.

\bibitem[\protect\citename{Maaten and Hinton}2008]{maaten2008visualizing}
Laurens van~der Maaten and Geoffrey Hinton.
\newblock 2008.
\newblock Visualizing data using t-sne.
\newblock {\em Journal of machine learning research}, 9(Nov):2579--2605.

\bibitem[\protect\citename{Majumder \bgroup et al.\egroup
  }2017]{majumder2017deep}
Navonil Majumder, Soujanya Poria, Alexander Gelbukh, and Erik Cambria.
\newblock 2017.
\newblock Deep learning-based document modeling for personality detection from
  text.
\newblock {\em IEEE Intelligent Systems}, 32(2):74--79.

\bibitem[\protect\citename{Matthews and
  Gilliland}1999]{matthews1999personality}
Gerald Matthews and Kirby Gilliland.
\newblock 1999.
\newblock The personality theories of hj eysenck and ja gray: A comparative
  review.
\newblock {\em Personality and Individual differences}, 26(4):583--626.

\bibitem[\protect\citename{Matthews \bgroup et al.\egroup
  }2003]{matthews2003personality}
Gerald Matthews, Ian~J Deary, and Martha~C Whiteman.
\newblock 2003.
\newblock {\em Personality traits}.
\newblock Cambridge University Press.

\bibitem[\protect\citename{Mikolov \bgroup et al.\egroup
  }2013]{mikolov2013distributed}
Tomas Mikolov, Ilya Sutskever, Kai Chen, Greg~S Corrado, and Jeff Dean.
\newblock 2013.
\newblock Distributed representations of words and phrases and their
  compositionality.
\newblock In {\em Advances in neural information processing systems}, pages
  3111--3119.

\bibitem[\protect\citename{Morin and Bengio}2005]{morin2005hierarchical}
Frederic Morin and Yoshua Bengio.
\newblock 2005.
\newblock Hierarchical probabilistic neural network language model.
\newblock In {\em Aistats}, volume~5, pages 246--252. Citeseer.

\bibitem[\protect\citename{Poria \bgroup et al.\egroup }2016]{poria2016deeper}
Soujanya Poria, Erik Cambria, Devamanyu Hazarika, and Prateek Vij.
\newblock 2016.
\newblock A deeper look into sarcastic tweets using deep convolutional neural
  networks.
\newblock {\em arXiv preprint arXiv:1610.08815}.

\bibitem[\protect\citename{Rajadesingan \bgroup et al.\egroup
  }2015]{rajadesingan2015sarcasm}
Ashwin Rajadesingan, Reza Zafarani, and Huan Liu.
\newblock 2015.
\newblock Sarcasm detection on twitter: A behavioral modeling approach.
\newblock In {\em Proceedings of the Eighth ACM International Conference on Web
  Search and Data Mining}, pages 97--106. ACM.

\bibitem[\protect\citename{Riloff \bgroup et al.\egroup
  }2013]{riloff2013sarcasm}
Ellen Riloff, Ashequl Qadir, Prafulla Surve, Lalindra De~Silva, Nathan Gilbert,
  and Ruihong Huang.
\newblock 2013.
\newblock Sarcasm as contrast between a positive sentiment and negative
  situation.
\newblock In {\em Proceedings of the 2013 Conference on Empirical Methods in
  Natural Language Processing}, pages 704--714.

\bibitem[\protect\citename{Stamatatos}2009]{stamatatos2009survey}
Efstathios Stamatatos.
\newblock 2009.
\newblock A survey of modern authorship attribution methods.
\newblock {\em Journal of the Association for Information Science and
  Technology}, 60(3):538--556.

\bibitem[\protect\citename{Tepperman \bgroup et al.\egroup
  }2006]{tepperman2006yeah}
Joseph Tepperman, David Traum, and Shrikanth Narayanan.
\newblock 2006.
\newblock " yeah right": Sarcasm recognition for spoken dialogue systems.
\newblock In {\em Ninth International Conference on Spoken Language
  Processing}.

\bibitem[\protect\citename{Tsur \bgroup et al.\egroup }2010]{tsur2010icwsm}
Oren Tsur, Dmitry Davidov, and Ari Rappoport.
\newblock 2010.
\newblock Icwsm-a great catchy name: Semi-supervised recognition of sarcastic
  sentences in online product reviews.
\newblock In {\em ICWSM}, pages 162--169.

\bibitem[\protect\citename{Wallace \bgroup et al.\egroup
  }2014]{wallace2014humans}
Byron~C Wallace, Laura Kertz, Eugene Charniak, et~al.
\newblock 2014.
\newblock Humans require context to infer ironic intent (so computers probably
  do, too).
\newblock In {\em Proceedings of the 52nd Annual Meeting of the Association for
  Computational Linguistics (Volume 2: Short Papers)}, volume~2, pages
  512--516.

\bibitem[\protect\citename{Wallace \bgroup et al.\egroup
  }2015]{wallace2015sparse}
Byron~C Wallace, Eugene Charniak, et~al.
\newblock 2015.
\newblock Sparse, contextually informed models for irony detection: Exploiting
  user communities, entities and sentiment.
\newblock In {\em Proceedings of the 53rd Annual Meeting of the Association for
  Computational Linguistics and the 7th International Joint Conference on
  Natural Language Processing (Volume 1: Long Papers)}, volume~1, pages
  1035--1044.

\bibitem[\protect\citename{Zhang \bgroup et al.\egroup }2016]{zhang2016tweet}
Meishan Zhang, Yue Zhang, and Guohong Fu.
\newblock 2016.
\newblock Tweet sarcasm detection using deep neural network.
\newblock In {\em Proceedings of COLING 2016, the 26th International Conference
  on Computational Linguistics: Technical Papers}, pages 2449--2460.

\end{thebibliography}

\end{document}